\documentclass[letterpaper, 10 pt, conference]{ieeeconf}  

\IEEEoverridecommandlockouts                              

\usepackage{graphics} 
\usepackage{epsfig} 
\usepackage{mathptmx} 
\usepackage{times} 
\usepackage{amsmath} 
\usepackage{amssymb}  
\usepackage{cite}
\usepackage{siunitx}
\usepackage{graphicx} 
\usepackage{epstopdf} 

\newcommand{\mZ}{\mathbb{Z}}

\newcommand{\cN}{\mathcal{N}}

\newcommand{\vc}{\textbf{c}}

\newcommand{\vx}{\textbf{x}}  
\newcommand{\vz}{\textbf{z}}   

\title{\LARGE \bf
UNIFY: Multi-Belief Bayesian Grid Framework \\ based on Automotive Radar
}

\author{Stefan Haag$^{1}$, Bharanidhar Duraisamy$^{1}$, Daniel Pfrommer$^{2}$, \\ 
Wolfgang Koch$^{3}$,  Martin Fritzsche$^{1}$ and J\"urgen Dickmann$^{1}$
\thanks{$^{1}$ Stefan Haag, Bharanidhar Duraisamy, Martin Fritzsche and J\"urgen Dickamnn are with Mercedes-Benz RD/ASF, Kolumbusstrase 21, 71063 Sindelfingen, Germany. E-mail: {\tt\small [firstname].[lastname]@daimler.com}}
\thanks{$^{2}$Daniel Pfrommer is with the University of Pennsylvania, Philadelphia, PA. E-mail: {\tt\small dpfrom@seas.upenn.edu}}
\thanks{$^{3}$Wolfgnag Koch is with Fraunhofer FKIE, Wachtberg, Germany. E-mail: {\tt\small [firstname].[lastname]@fkie.fraunhofer.de}}    
}

\begin{document}

\maketitle
\thispagestyle{empty}
\pagestyle{empty}

\begin{abstract}
Grid maps are widely established for the representation of static objects in robotics and automotive applications. Though, incorporating velocity information is still widely examined because of the increased complexity of dynamic grids concerning both velocity measurement models for radar sensors and the representation of velocity in a grid framework. 
In this paper, both issues are addressed: sensor models and an efficient grid framework, which are required to ensure efficient and robust environment perception with radar.
To that, we introduce new inverse radar sensor models covering radar sensor artifacts such as measurement ambiguities to integrate automotive radar sensors for improved velocity estimation. 

Furthermore, we introduce UNIFY, a multiple belief Bayesian grid map framework for static occupancy and velocity estimation with independent layers. 
The proposed UNIFY framework utilizes a grid-cell-based layer to provide occupancy information and a particle-based velocity layer for motion state estimation in an autonomous vehicle's environment.
Each UNIFY layer allows individual execution as well as simultaneous execution of both layers for optimal adaption to varying environments in autonomous driving applications.  

UNIFY was tested and evaluated in terms of plausibility and efficiency on a large real-world radar data-set in challenging traffic scenarios covering different densities in urban and rural sceneries.
\end{abstract}

\section{Introduction}

Developing perception systems in automated driving for highly dynamic and unstructured environments is a significant field of current research. 
Environment perception frameworks for an autonomous driving application must be robust to a wide range of scenarios, such as dense high-speed motorway scenarios or inner-city scenarios on the one hand and less dense suburban scenarios or scenarios in rural environments on the other hand. 
Thus, the perception system has to fulfill a diverse set of requirements.  
For instance, long-range obstacle detection may be necessary on a high-speed motorway, whereas medium-range detection would be sufficient in low-speed urban scenarios. 
Any such framework must be capable of processing large volumes of data from a heterogeneous set of sensors, and so must have a generic world model that can incorporate new sources of information. 

\begin{figure}[htb]
    \centering
    \includegraphics[scale=1,trim=95 0 85 0,clip]{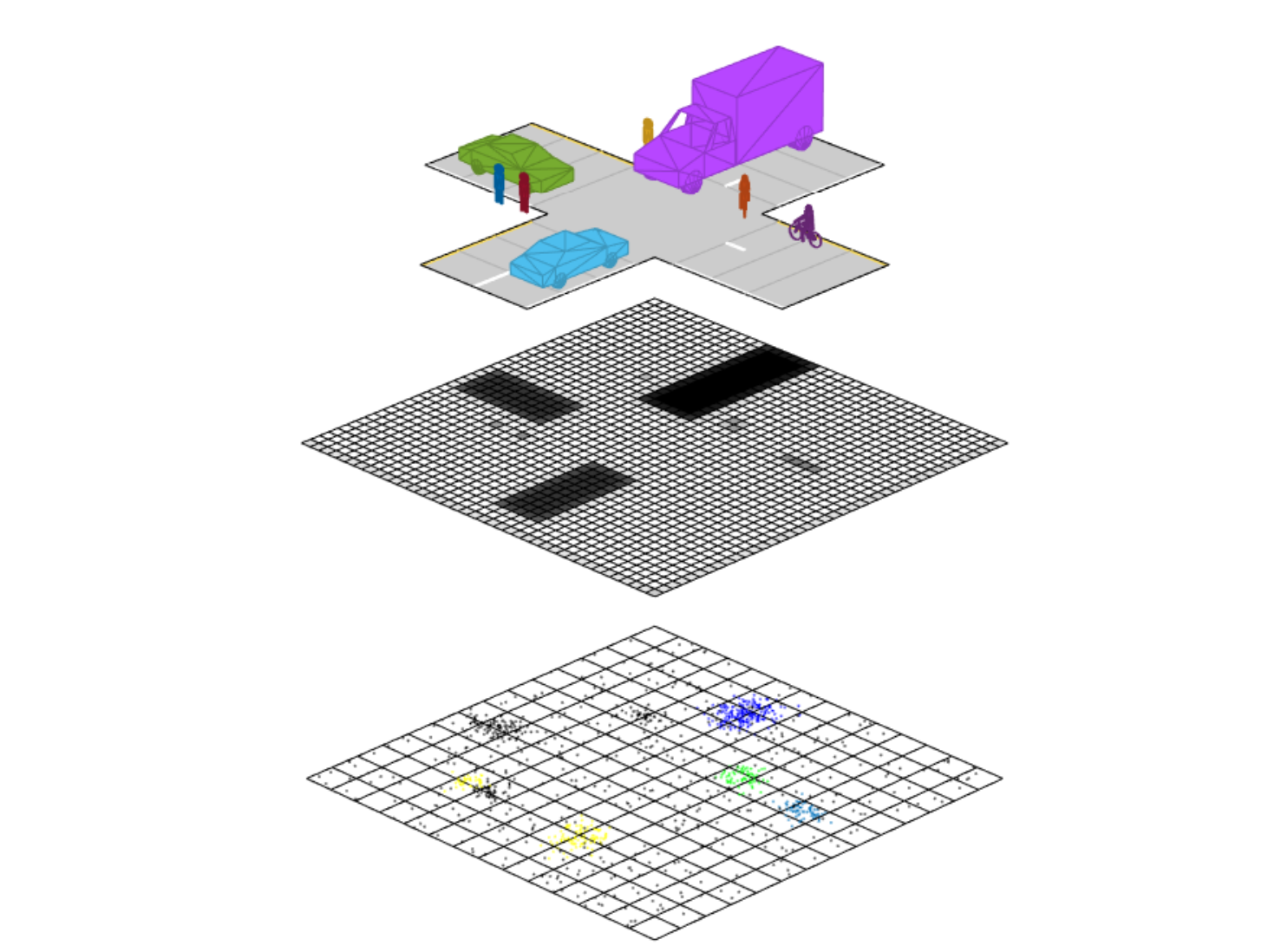}
    \caption{Expected UNIFY perception results independent of utilized sensors of an example traffic scenario at an intersection. The top frame indicates the real world representation. The middle frame indicates the occupancy layer, here in binary Bayes form. Grey tones indicate cell occupancy probability independent of object dynamics. The lower frame shows the particles which represent the dynamic information of the given scenario in the independent velocity grid. Black particles indicate static objects}
    \label{fig:layer}
\end{figure}

Rather than pursuing clustering and object detection techniques, we focus on generic environment mapping methods. 
The most common approach in this domain is the static occupancy grid map, which models occupied and free space for static or slow-moving targets. 
This information is required in automotive applications for trajectory planning, collision avoidance and self-localization. 
However, static maps cannot handle dynamic obstacles common to autonomous driving, such as traffic and pedestrians. 
One possible approach to integrate dynamic object tracking is to use a combination of clustering and hand-crafted filtering algorithms such as Extended Object Tracking (EOT) \cite{Haag2018a}, where each object is explicitly instantiated.
While these systems are effective, they are not flexible in size and shape of the tracked objects, meaning they do not generalize sufficiently and have numerous complex edge cases. 
Complexity grows enormously fast for an increasing number of objects.

This work will naturally integrate dynamic object tracking into a static model by augmenting the grid map to include velocity information as part of a dynamic occupancy grid map. 
This approach is highly computationally expensive when compared to standard object tracking techniques. 
However, with recent developments in high-performance mobile computing platforms, dynamic occupancy grid maps have garnered increased attention due to their highly flexible object representation and inherent scalability to more complex environments and sensor models.

For this reason, novel radar sensor models are required to cover ambiguities and to improve reliability on the whole perception system. 
Nevertheless, this work is kept modular in an independent layer structure, which allows the integration of additional sensors and information, for example, lidar sensors for height information or camera sensors for classification information.

In the following paper, we present UNIFY, a novel framework for integrating point cloud data from heterogeneous sensors in a unified manner. 
The system augments an occupancy grid map utilizing either Binary Bayesian inference or Dempster-Shafer belief masses 
with a particle-based system for velocity inference for simultaneous and independent occupancy and velocity estimation as it is shown in Fig. \ref{fig:layer}. Occupancy and velocity are tracked in mostly independent layers. The figure shows the projection of the real world representation in the top layer to the occuancy layer in the middle and the velocity layer on the bottom. 
The structure and implementation of UNIFY was designed from the ground up for real-time execution on modern hardware, utilizing General Purpose Graphics Processing Unit (GPGPU) toolkits as part of a high-performance C++ codebase. 
The result is a pluggable, extensible framework for occupancy grid computations.

The paper is structured as follows. Sec. \ref{sec:RW} discusses related work concerning radar sensor models and occupancy grid maps.
Radar specific measurement models are introduced in Sec \ref{sec:RMM}.
The proposed UNFIY framework with its new developed layer structure is introduced in Sec. \ref{sec:UNIFY}
Experimental results are shown in Sec. \ref{sec:ER}.
Sec. \ref{sec:conclusion} concludes the paper.

\section{Related Work}
\label{sec:RW}

Static occupancy grids are widely used in robotics \cite{Thrun2005} because they allow high performing implementation in logarithmic representation and fixed memory usage.   
Often, static occupancy grids are applied in combination with self-localization problems. 
However, ego-position and ego-motion information are mostly available in automotive applications \cite{Yuan2015a} and free space has to be determined for further applications \cite{Mouhagir2018} which requires handling of both static and dynamic targets.
For that reason, the major challenge of applying occupancy grid maps in automotive applications lies in efficiently integrating velocity information into a grid-based framework. 

Several approaches exist to represent velocity information of grid cells, which is significantly more challenging and complex than occupation information because the state space can not be discrete and the distribution can be diffuse if several different objects are covering neighboring cells.  
In the first approach from \cite{Coue2003}, a static grid was augmented by adding a histogram to each grid cell to cover the velocity distribution of an object hypothetically covering the cell. 
With this approach, a constant amount of memory is required independent of the occupation state of the cell.
Recent works \cite{Danescu2014, Nuss2015a, Tanzmeister2017} have utilized particle systems in combination with radar measurements to tractably model and measure cell velocity.  
Many prior approaches, such as \cite{Tanzmeister2017}, provide velocity integration with distinct dynamic and static occupancy belief estimators for each cell.
For instance, \cite{Nuss2015a} introduces a particle-based grid approach where radar range-rate measurements are integrated to update particle weights. 
Radar sensors are essential for dynamic grids due to their ability to measure the radial velocity, unlike all other sensors directly.

Besides the representation, suitable (inverse) sensor models are essential to obtain accurate and reliable occupancy and velocity estimators. 
Radar sensors are harder to model since they differ significantly from ideal models in practice \cite{Werber2015, Haag2018} due to several radar specific artifacts such as multi-path propagation, transvision and ambiguous measurements that apply to automotive radars as well.

\section{Radar Sensor Models}
\label{sec:RMM}

Proper measurement models are crucial for obtaining high-quality results when using Bayesian Occupancy grids.
The measurement model has to adjust to every sensor and sensor type differently for optimal tracking and fusion results.
In this section, a novel inverse radar sensor model is introduced for application in the UNIFY framework.

Radar sensors provide point clouds of a varying number of detections with every scan.
Each detection provides position information as range and azimuth measurements of detected obstacles along with the radial velocity and the corresponding measurement noises.
Often, measurement errors are assumed to be zero mean Gaussian distributed. 
However, measurement ambiguity errors can not be resolved entirely by the sensor's signal processing logic. Thus,  incorrect locations or velocities are be obtained regularly where the measurement errors exceed the provided measurement noise by a significant magnitude.
The proposed models allow consideration of ambiguities in terms of position and velocity measurements that can not be covered from a Gaussian distribution because of the appearance of several peaks in the likelihood function.

We assume that position and measurements are uncorrelated. 
Hence, the measurement model for the position and velocity update for the particle update in UNIFY can be divided into the occupancy (\ref{eq:mo}) likelihood and the velocity likelihood (\ref{eq:vm}). 

\begin{align}
\label{eq:ParticleMeasModel}
\begin{aligned}
    \Lambda^{i,j} & = \Lambda \left(  \vz_k^j \middle\vert \vx_{k-1 \vert k}^i \right) \\ 
    & = \Lambda_O \left(  \vz_k^j \middle\vert \vx_{k-1 \vert k}^i \right) \cdot \Lambda_V \left(  \vz_k^j \middle\vert \vx_{k-1 \vert k}^i \right)
\end{aligned}
\end{align}

In case of static-only Dempster-Shafer belief estimation, measurement estimators for occupancy mass and free mass have to be modeled depending only on position measurements.
But the range-rate measurements allow distinction between static and dynamic occupancy which is only possible with radar sensors so that a grid representing only static targets are possible.

\subsection{Position Measurement Model}

\begin{figure}[thb]
    \centering
    \includegraphics[width=\columnwidth]{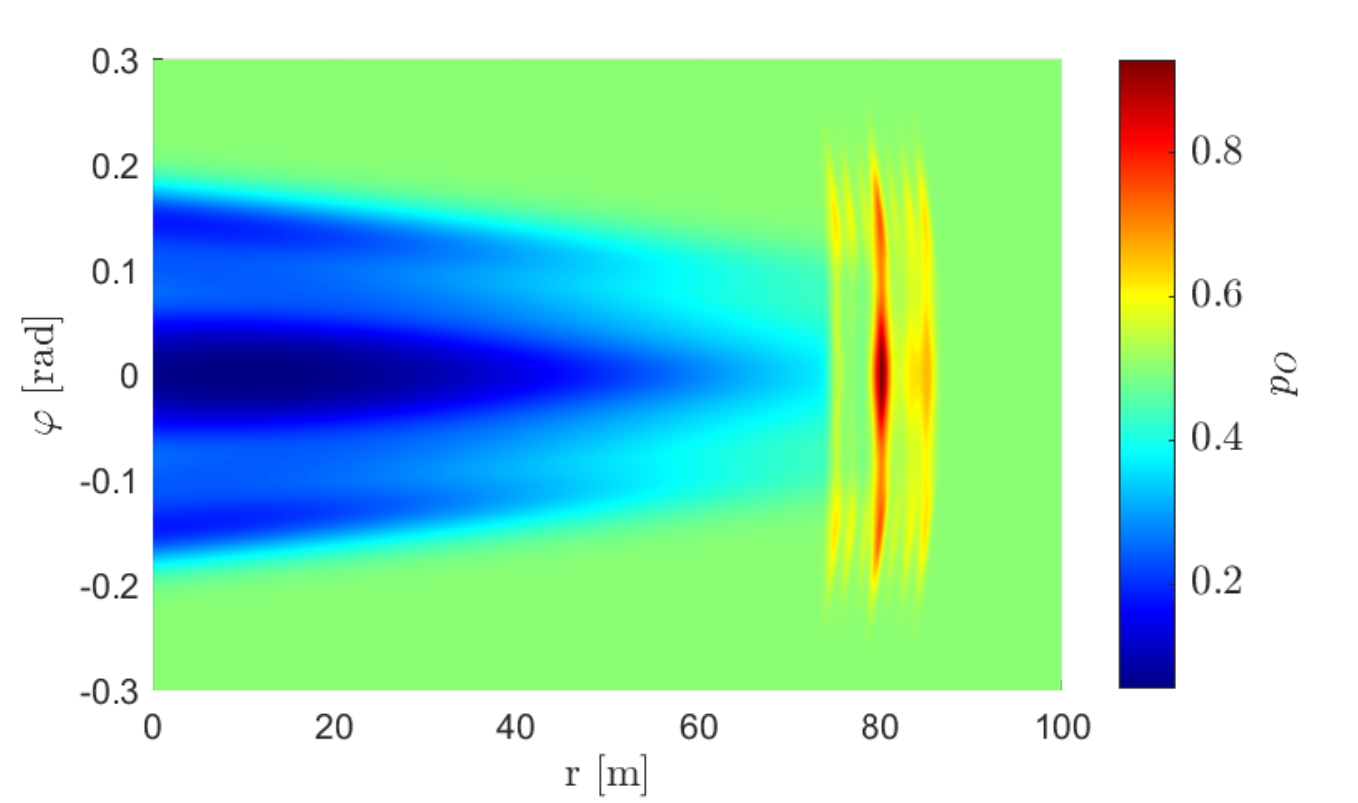}
    \caption{Radar occupancy probability for measurement at $z = \left(80, 0 \right)$, likely free space is colored in blue colors, likely occupied space is colored in yellow to red. Unsolvable space is colored in green. The density probability is maximal at the actual measurement position but shifted peaks provoked from ambiguities are visible. Free space is visible in front of occupied cells}
    \label{fig:po}
\end{figure}

Concerning position measurements, we assume that the occupancy likelihood is higher the closer a measurement is located to the cell. 
With regard to the measurement ambiguities, an obtained measurement could be incorrect by integer multiples of some $\Delta_r$ in the radial distance and $\Delta_\varphi$ azimuth directions, for instance, if a ray gets deflected from external influences and hits another cell in the antenna array.
Therefore, the shift sizes $\Delta_r$ and $\Delta_\varphi$ are sensor-specific parameters depending on the sensor's antenna resolution. 
The integer shifts, $i\Delta_r$ and $j\Delta_\varphi$ are modelled as $i, j$, are sampled from independent binomial distributions with binomial parameter $p = 0.5$ on some fixed index sets $I = J = \{-k, ..., k\} \subset \mZ$. 
In our case with $k = 1$.
The probability that a specific pair $(i, j)$ is correct can be derived as $\phi_{i,j}$ where $\sum_{i \in I} \sum_{j \in J} \phi_{i,j} = 1$.
Furthermore, we assume that besides the wrong ambiguity resolution, Gaussian measurement noise is still present.
Therefore, the measured occupancy likelihood for cell $c$ and measurement $z = \left(r,\varphi \right)$ is modeled as
\begin{align}
\label{eq:mo}
\begin{aligned}
    \Lambda_O^c \left( \vc \vert \vz\right) & = \eta_O \sum_{i \in I} \sum_{j \in J} \phi_{i,j} \\ & \cdot \cN \left( P_{i,j}  ; \begin{pmatrix} r \\ \varphi \end{pmatrix}, \begin{pmatrix} \sigma_r^2 & \rho \sigma_r \sigma_\varphi \\ \rho \sigma_r \sigma_\varphi & \sigma_\varphi^2 \end{pmatrix} \right)
\end{aligned}
\end{align}
with the hypothetical cell location, shifted by discrete steps
\begin{align}
    P^c_{i,j} = \begin{pmatrix} r_c + i \Delta_r \\ \varphi_c + j \Delta_\varphi \end{pmatrix}
\end{align}
and constant factor $\eta_O > 0$, measurement noise $\left( \sigma_r^2, \sigma_\varphi^2 \right)$, and correlation $\rho = \text{sign} (\varphi_c + j \Delta_j) \cdot \rho_0$, $\rho_0 \in (0,1)$.

Under consideration of equal assumptions, the measurement ambiguities errors are integrated in the same manner in the free space likelihood model.
\begin{align}
\label{eq:mf}
    \Lambda^c_F \left( \vc \vert \vz \right) = \eta_F \sum_{i \in I} \sum_{j \in J} \phi_{i,j} \left\{ \begin{matrix} 0, \ \text{ if } r \ge r_c + i \Delta_r \\  
    \cN \left( P^c_{i,j}; \mu_F^c,  \Sigma_F^c  \right), \ \text{else} \end{matrix} \right. 
\end{align}

With mean and covariance 
\begin{align}
    \mu_F^c = \begin{pmatrix} 0 \\ \varphi \end{pmatrix}
    \text{ and } 
    \Sigma_F^c = \begin{pmatrix}  \gamma^2 r^2 & \rho \gamma r \\ \rho \gamma r & \sigma_\varphi^2 \end{pmatrix}
\end{align}
Thereby, the occupancy probability is then obtained with 
\begin{align}
    p_O \left( \vc \vert \vz \right) = \frac12 \left( 1 + m_O\left( \vc \vert \vz \right) - m_F\left( \vc \vert \vz \right)  \right)
\end{align}

An example for the new occupancy probability considering position measurement ambiguities for the Binary Bayes update as combination of (\ref{eq:mo}) and (\ref{eq:mf}) is shown Fig. \ref{fig:po}.
In Contrast to the ideal measurement model, the occupied space is further distributed and not centered around a single peak. Thought the highest peak is still located at the object position. Further peaks are distributed in the fixed distance to each other in both dimensions.
Blue and green colored areas indicate likely free space in between sensor and object.
It is divided into several valleys instead of the single valley, provided from the standard model.
An impact on both states occupied, and free is visible.
An ideal measurement model applicable to lidar measurements would be obtained with $I = J = \{ 0 \}$. 

\subsection{Velocity Measurement Model}

\begin{figure}[thb]
    \centering
    \includegraphics[width=\columnwidth]{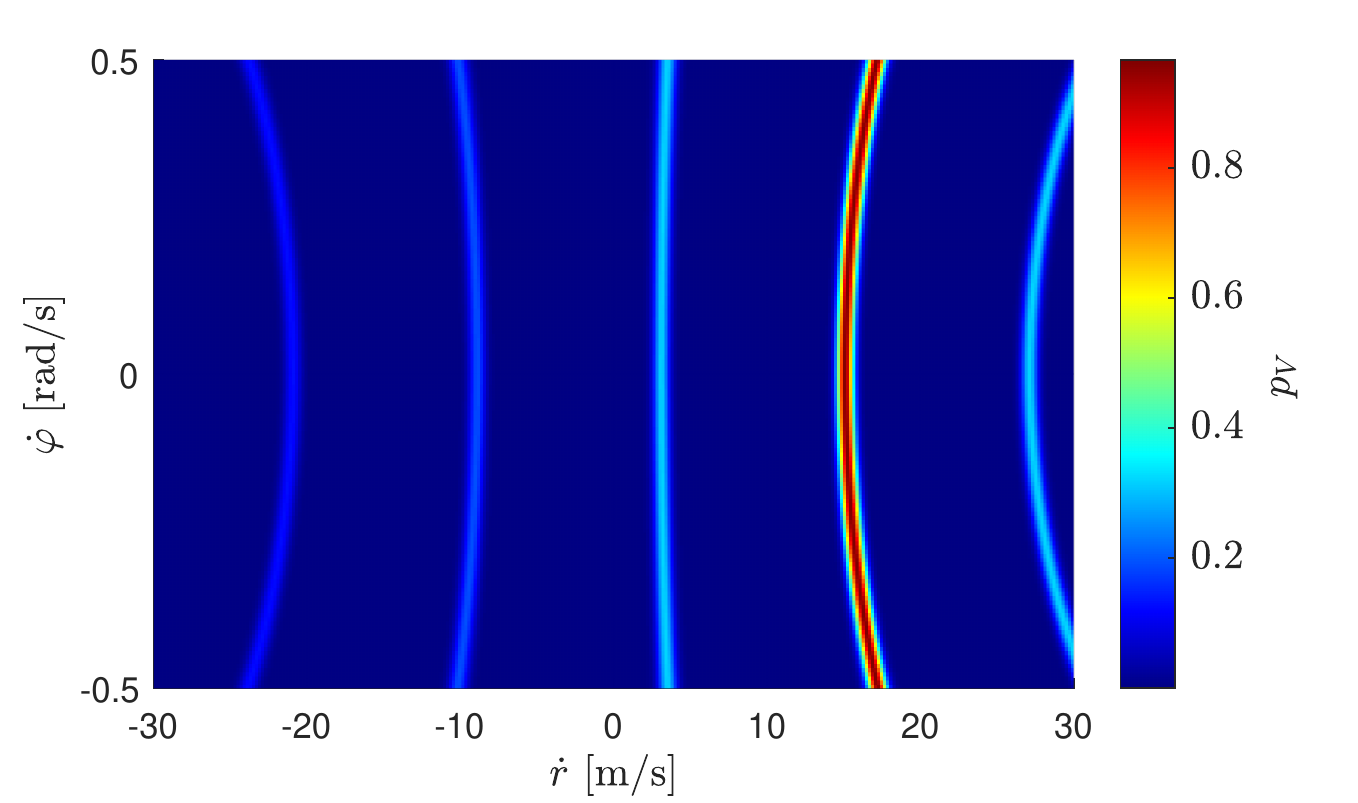}
    \caption{Radar velocity probability for the cell located at $(80,0)$ with measurement $z = \left(80, 0, 15 \right)$, the sensor is located at the origin. Cross velocity is not resolvable which leads to lines instead of peaks in the probability plot. The model leads to several distinct velocity hypothesis with potentially large difference}
    \label{fig:pv}
\end{figure}

\begin{figure*}[htb]
\centering
    \includegraphics[width=\textwidth]{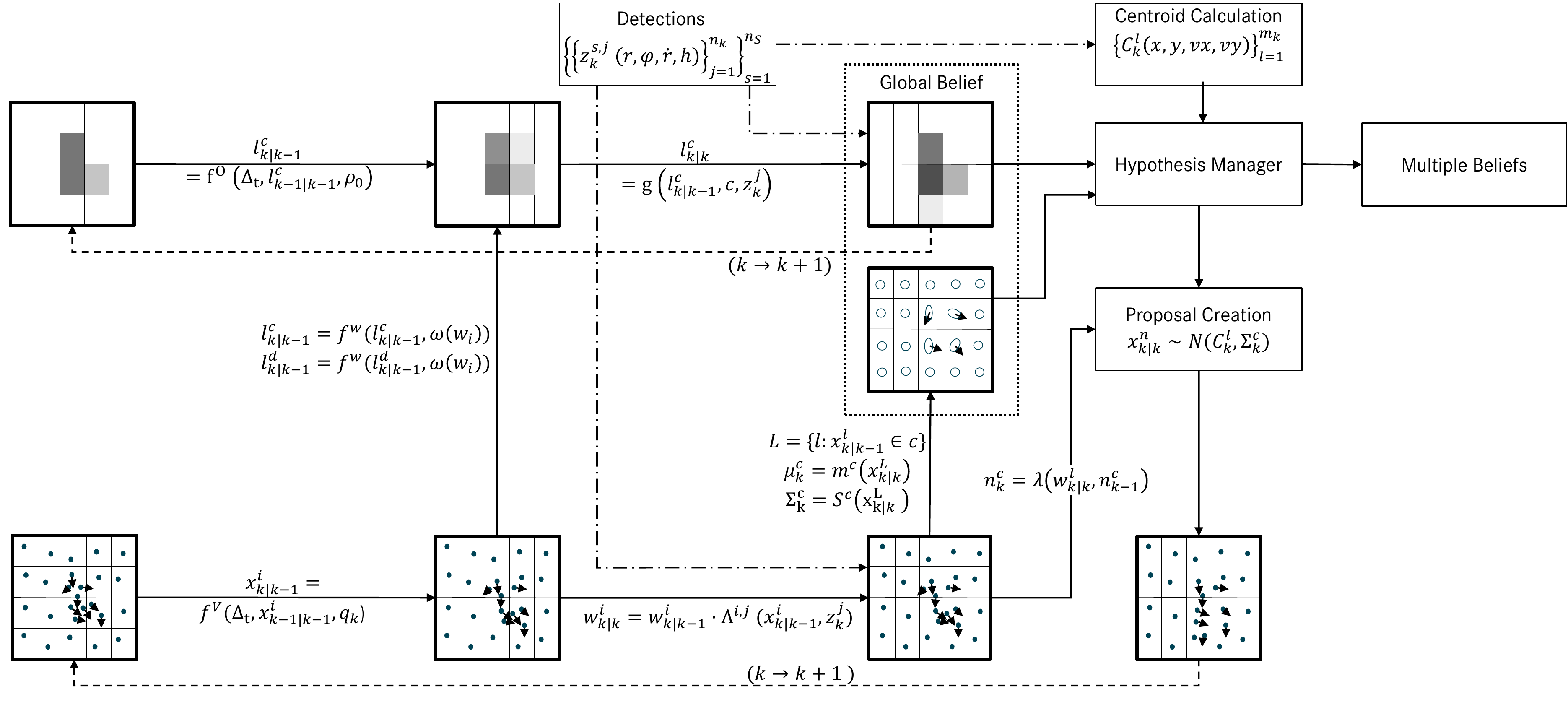}
    \caption{UNIFY process cycle. 
    The structure is explained in detail in Sec. \ref{sec:UNIFY}. The grid images colored in black and white in the upper row represent the occupancy layer, the grid images in the lower column represent the velocity layer along with its particles shown as dots with arrows to display their velocity. The grid in the middle row indicates the velocity belief for each cell represented as Gaussian for further application in the Hypotheses Manager}
    \label{fig:UnifyFlow}
\end{figure*}

A likelihood model including the radial velocity is required to exploit radar range-rate measurements. 
It is essential to provide accurate velocity estimation in our velocity layer because it is required for particle creation and the particle weight update.

Range-rate measurements are potentially shifted by an integer multiple of some interval $\Delta_{\dot{r}}$ that is sensor-specific and with a given shift index $l \in L \subset \mZ$. 
The particle update $\Lambda^{i,j}$ is calculated with the nearest measurement $\vz_k^j$ so that the velocity measurement model yields in a Gaussian mixture model, too.
\begin{align}
\label{eq:vm}
    p_V \left(  \vz_k^j \middle\vert \vx_{k-1 \vert k}^i \right) = \eta_V \cdot \sum_{l \in L} \phi_l \cN \left( \dot{r}_k^j; h_{\dot{r}} \left( \vx_{k \vert k-1}^i, \sigma^2_{\dot{r}}  \right) \right)
\end{align}
With the Doppler measurement model $h_{\dot{r}} \left( \cdot \right)$ \cite{Haag2019a}.

Fig. \ref{fig:pv} shows the proposed measurement model for a measured cell the intensity is plotted over the velocity distribution of a measured cell. It shows that the difference between each velocity hypothesis can be very large so that the inclusion of this type of errors in the velocity measurement model is crucial. 
Because of the large distances between each velocity hypothesis a multi hypothesis multi-hypothesis tracking for example with a particle filter, is highly recommended to cover all distinct hypotheses and benefit from the new model.  

\section{UNIFY}
\label{sec:UNIFY}

UNIFY provides simultaneous occupancy and velocity estimation which shall be estimated independently.
Therefore, the UNIFY framework consists of an occupancy layer and a velocity layer that are stored independently and separately. 
Each layer is represented as grid of cells which can be configured completely independent of the other layer in terms of size and resolution.
Fig. \ref{fig:UnifyFlow} shows an update cycle of both layers in simultaneous execution. 
We assume each occupancy grid cell to be probabilistically independent of all other cells. 
This is a common assumption in occupancy grid maps, which, although clearly false, is necessary in order to compute the posterior and prior occupancy probabilities in a feasible manner.

In particular, cells in the occupancy layer are updated independently of cells in the velocity layer. 
Both layers are Bayesian filters where each update step can be divided into time update (prediction) and a measurement update (correction). 
Fig. \ref{fig:UnifyFlow} shows a full iteration of the algorithm.

The occupancy layer probability distribution is modelled directly, whereas the velocity layer utilizes a particle system which provide a discrete but highly flexible approximation of the cell's velocity state. 
Statistics such as mean velocity and variance are computed for each cell from this particle system to provide the global belief for integration into higher-level algorithms.

\subsection{Grid Adjustment}

The grid is kept in a global coordinate system but it is only populated around the sensor platform (SP), where measurements are available. 
For the provided measurement set up shown in \ref{fig:FOV}, a grid extending \SI{75}{\metre} in front of the vehicle and \SI{75}{\metre} behind the vehicle was chosen. The grid has \SI{150}{\metre} of space to the left and right of the car. 
Since the grid is kept in a global coordinate system in which the ego-vehicle moves, the vehicle is simply translated within the grid on movement. 
The grid does not need to be transformed on vehicle rotation as it is globally aligned, eliminating an expensive and lossy rotation step. 
All sensor information has to be transformed into the global reference frame, before being fused into the grid.

As the vehicle moves, the grid cells behind the vehicle are discarded behind the SP and new cells initialized in front of the SP from time to time if the SP has traveled a certain distance.
Thereby, the overall grid size is kept constant and the populated areas are always located around the sensor platform but information of traveled areas is deleted.
Thereby the grid can be adjusted to changing environments in terms of grid size and cell size.

\subsection{Occupancy Layer}

As previously mentioned, the occupancy layer grid representation is configurable. 
The occupancy cell state can be tracked with a Binary Bayesian (BB) approach where solely the logarithmic occupancy probability $l_k^c = \log(p_k^O)$ of each cell is estimated. 
Alternatively, a Dempster Shafer (DS) theory of evidence model can be used \cite{Mouhagir2018a}. 
In the DS model, occupancy mass $M_O^c$ and free mass $M_F^c$ are estimated simultaneously so that unresolved cell with conflicting information can be distinguished from unobserved cells, where no information is available. Both state representations are supported in UNIFY.

The occupancy layer invokes a prediction and an update step as shown in \ref{fig:UnifyFlow}, similar to static grids. 
Although not technically static, updates are made to the occupancy layer as if all measurements belonged to static objects. 
The dynamics information of objects in the occupancy layer is tracked in the velocity layer, which adjusts the occupancy layer accordingly during the mass transfer step between iterations. 
Particles carry occupancy belief masses with respect to their weight $\omega(w_i)$ when they move from one occupancy grid cell to another during the particle propagation phase. 
This is the only point where the velocity layer influences the occupancy layer. 
Without this step, the occupancy layer would be identical to a static grid.

\subsection{Velocity Layer}

The velocity layer shown in the lower part in Fig. \ref{fig:UnifyFlow} is kept completely independent of the occupancy layer so that tracking of dynamic objects can vary according to varying challenges. 
Extended Object Tracking \cite{Haag2019a} could be performed with less computational effort and memory in less dense scenarios. 
But a fixed size grid has fixed memory and effort requirements.

The velocity layer divides the SP's environment into grid cells that are not necessarily equal to the occupancy grid cells.
Each grid cell is populated with particles to model the velocity distribution. 
The particles consist of position, velocity parameters and a weight parameter.
\begin{align}
    \vx_k^l = \left( x_k^l, y_k^l, \dot{x}_k^l, \dot{y}_k^l, w_k^l \right)^T
\end{align}

In the prediction step, the particles are propagated according to the motion model $f^V$ and system noise vector $q_k$.
Each cell has an array of the particles contained within that cell for quick access to each particle. 
The array must be kept up-to-date after every prediction. 
Particles that cross cell borders are matched with their new cells in the propagation step. 

The particle update is performed in several steps. 
First the $n$ closest measurements are found for each cell and for each sensor. 
Second, out of the matched measurements the optimal measurement is found for each particle using the sensor model with which the particle weight-update (\ref{eq:ParticleMeasModel}) is performed. 
Third, After all particle weights are updated, weighted mean $\mu_k^c$ and covariance estimators $\Sigma_k^c$ are calculated for each grid cell to obtain estimators for the global belief.

Particle resampling is performed in every cycle.
The new number of particles in the cell $n_k^c$ is calculated based both on the old number of particles and the average weight.
Occupied cells are populated more densely, giving higher resolution in the estimated velocity distribution for moving objects.
The number of particles in a cell is limited with a lower and an upper bound so that velocity information is available for every grid cell and upper and lower bounds for the storage usage and computational effort can be derived.

Velocity cells are initialized with uniformly distributed particles covering a small range of velocities to keep up with slowly moving targets.
New cells are initialized to have higher initial velocities if there are high-speed radar measurements nearby.
For the purpose of particle initialization, dynamic radar measurements are clustered and associated to near grid cells in the hypotheses manager shown in Fig. \ref{fig:UnifyFlow}. 
If a cluster is associated to a cell, new particles are drawn with the cluster velocity mean transformed into Cartesian space.


\section{Experimental Results}
\label{sec:ER}

\begin{figure}[htb]
    \centering
    \includegraphics[width=\columnwidth]{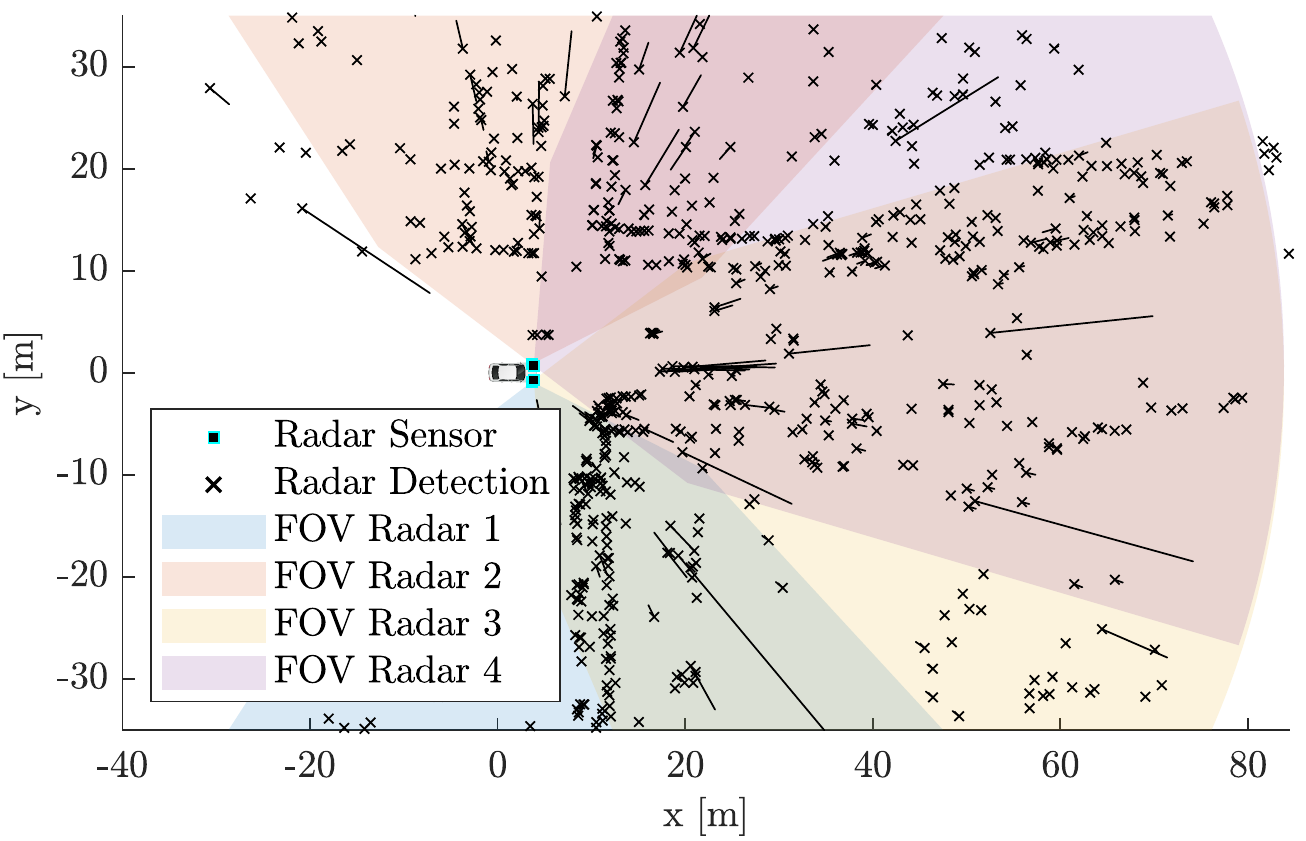}
    \includegraphics[width=\columnwidth]{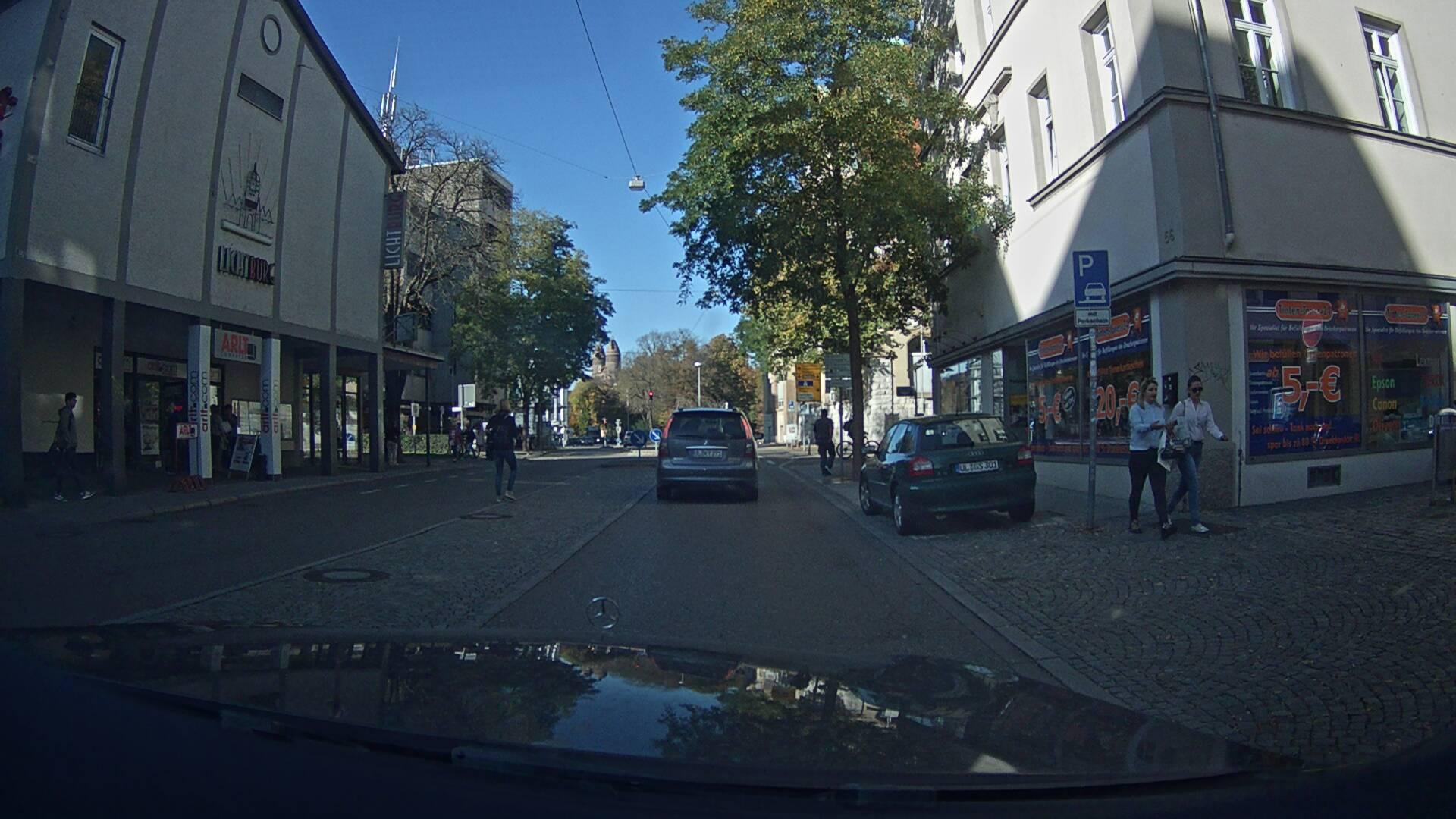}
    \caption{Utilized experimental Sensor Platform (SP) equipped with four automotive radar sensors with overlapping fields of view in image above. Position and range-measurements from an urban scenario shown in the image below}
    \label{fig:FOV}
\end{figure}

\begin{figure*}[htb]
    \centering
    \includegraphics[width=\textwidth]{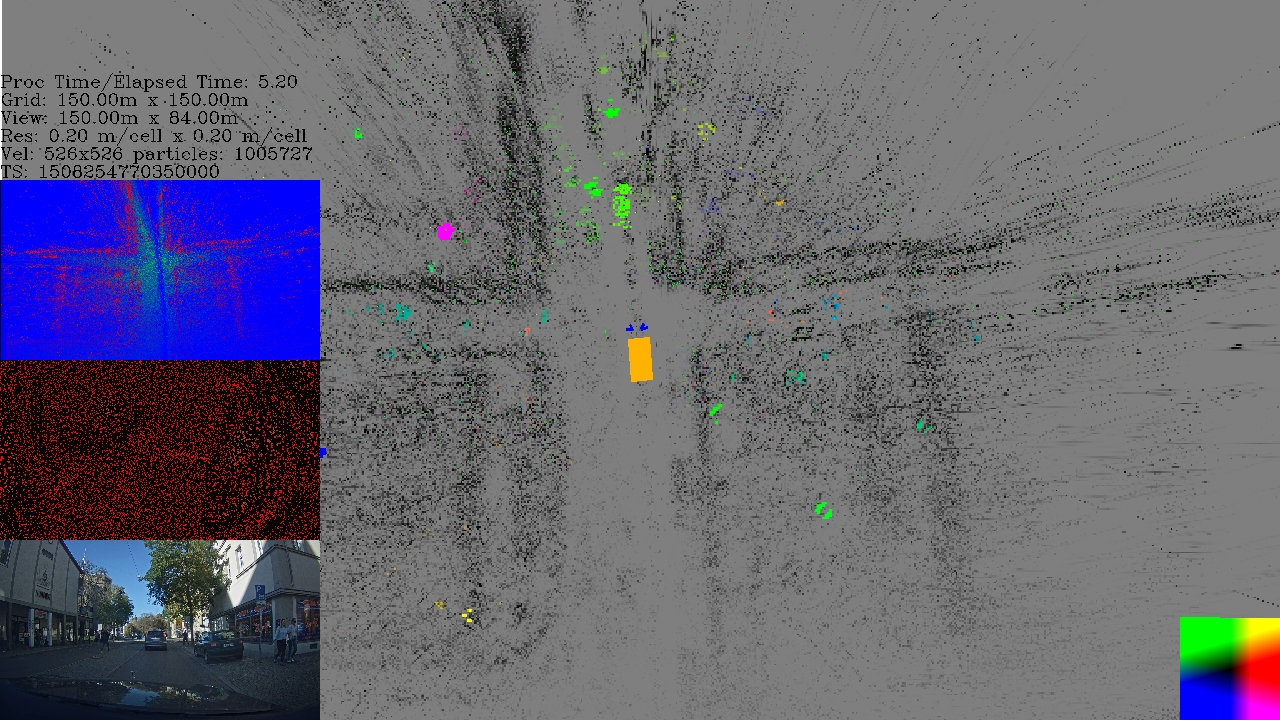}
    \caption{UNIFY in urban scenario shown in Fig. \ref{fig:FOV} the main view shows static occupancy probabilities in grey colours. Dynamic objects are coloured according to the velocity legend in the lower right corner which appear as several colored clusters. The observed car in front of the ego vehicle is visible as green block, moving in the same direction as the SP. In the views on the left, from top to bottom: grid statistics such as processing time compared to recording time, the Dempster Shafer grid with occupied space in red and free space in green, the middle view show particle density where red areas indicate a high particle density and black low particle density the lower view shows the dash cam image of the current scenario}
    \label{fig:grid}
\end{figure*}

We had access to an extensive radar data set recorded with the sensor set-up shown in Fig. \ref{fig:FOV}.
The data set covered highways scenarios, city scenarios, and rural scenarios. 
Urban scenarios were the most challenging for UNIFY because of the high density of dynamic objects near the SP, providing much dynamic radar measurement. 
Thus, a massive amount of particles would be required for each cell to cover the multitude of different velocity hypotheses, which would increase the computational effort significantly.

An example of simultaneous occupancy and velocity tracking with UNIFY is shown in Fig. \ref{fig:grid}.
The SP is shown as a yellow rectangle in the middle of the main view. The road boundaries (i.e., curbside), buildings, and stationary vehicles are visible in the visible part in front and the previously covered area behind the vehicle. The curbs give only conflicting measurements without height information, as shown in the Dempster-Shafer plot in the top left, where green is free mass, and red indicates occupied mass. 
Due to some static occupied mass in front of the SP, the free space is mostly detected on the sides.

House walls are visible as distinct black lines in the static layer. Furthermore, the curbstones between the road and sidewalks are detected so that the road structure is apparent, for example, with the intersections visible behind and next to the ego vehicle. 
Those results for a static environment and the modular structure of UNIFY would allow the integration of road map information for precise self-location and scene understanding, which is essential for further applications such as trajectory planning tasks. 

Clusters of dynamic measurements are used to create fast-moving particles. 
This allows us to populate unresolved cells only with static and slow-moving particles so that a lot of particles and effort can be saved to keep UNIFY traceably. 

The car in front of the SP is visible as a green cluster in the vehicle lane. The color corresponds to the mean velocity of the associated particles. 
Pedestrians are vaguely visible as they move slower and have fewer associated radar measurements than cars.
Some false positives are visible such as the pink cluster, which is due to false range-rate measurements. 
However, these false positives are suppressed quickly in the velocity layer when no further such measurements appear.
Mirror objects due to multi-path bounces cannot be suppressed when only radar data is available. 
The dynamic layer can be exchanged with a standard EOT approach for dynamic objects in less dense scenarios. Instead of particles, the objects would move cell belief weights. Furthermore, classification information is required for better scene awareness.

\section{Conclusion}
\label{sec:conclusion}

This paper introduced UNIFY, a multi-belief Bayesian grid map framework providing static and dynamic occupancy estimation with automotive radar.
The framework is structured in independent layers for a dynamic adaption to varying traffic scenarios. 
We introduced new radar sensor models covering measurement ambiguities for application in our grid map framework. 
Our proposed models differ from standard single-Gaussian-based models, but our models cover more complex conditions and provoke a broader hypothesis spread.
 The proposed Gaussian mixture form allows fast and efficient updates as Gaussian mixtures. 

We have shown the system's real-world feasibility in complex and challenging urban environments by applying UNIFY an extensive real-world radar data set. 
Future work involves the integration of additional information for more efficient and robust results and better context-awareness. 

\section*{Acknowledgment}

A part of this work is funded by German Federal Ministry of Education and Research (BMBF) and Federal Ministry for Economic Affairs and Energy (BMWi) through Radar4FAD project with the project grant no. 16ES0560.
	
\bibliography{library} 
\bibliographystyle{ieeetr}

\end{document}